\documentclass[10pt,twocolumn,letterpaper]{article}

\usepackage[pagenumbers]{cvpr} 

\usepackage{bm}
\usepackage{color}
\usepackage{graphicx}
\usepackage{amsmath}
\usepackage{amssymb}
\usepackage{booktabs}
\usepackage{xcolor}
\usepackage{multirow}
\usepackage[normalem]{ulem}
\newcommand\hl{\bgroup\markoverwith
  {\textcolor{yellow}{\rule[-.5ex]{2pt}{2.5ex}}}\ULon}

\usepackage[pagebackref,breaklinks,colorlinks]{hyperref}

\usepackage[capitalize]{cleveref}
\crefname{section}{Sec.}{Secs.}
\Crefname{section}{Section}{Sections}
\Crefname{table}{Table}{Tables}
\crefname{table}{Tab.}{Tabs.}

\def\cvprPaperID{5789} 
\def\confName{CVPR}
\def\confYear{2022}

\begin{document}

\title{Adversarial Attack by Shadow}
\title{Shadows can be Dangerous: Stealthy and Effective Physical-world Adversarial Attack by Natural Phenomenon}
\author{Yiqi Zhong$^{1}$, Xianming Liu$^{1,2}$\thanks{Corresponding author}, Deming Zhai$^{1}$, Junjun Jiang$^{1,2}$, Xiangyang Ji$^{3}$\\
$^{1}$Harbin Institute of Technology,
$^{2}$Peng Cheng Laboratory,
$^{3}$Tsinghua University\\
{\tt\small 21s003117@stu.hit.edu.cn, \{csxm, zhaideming, jiangjunjun\}@hit.edu.cn, xyji@tsinghua.edu.cn}
}
\maketitle

\begin{abstract}
 
 Estimating the risk level of adversarial examples is essential for safely deploying machine learning models in the real world. One popular approach for physical-world attacks is to adopt the “sticker-pasting” strategy, which however suffers from some limitations, including difficulties in access to the target or printing by valid colors. A new type of non-invasive attacks emerged recently, which attempt to cast perturbation onto the target by optics based tools, such as laser beam and projector. However, the added optical patterns are artificial but not natural. Thus, they are still conspicuous and attention-grabbed, and can be easily noticed by humans.
In this paper, we study a new type of optical adversarial examples, in which the perturbations are generated by a very common natural phenomenon, \textit{shadow}, to achieve naturalistic and stealthy physical-world adversarial attack under the black-box setting. We extensively evaluate the effectiveness of this new attack on both simulated and real-world environments. Experimental results on traffic sign recognition demonstrate that our algorithm can generate adversarial examples effectively, reaching 98.23\% and 90.47\% success rates on LISA and GTSRB test sets respectively, while continuously misleading a moving camera over 95\% of the time in real-world scenarios. We also offer discussions about the limitations and the defense mechanism of this attack\footnote{Our code is available at \href{https://github.com/hncszyq/ShadowAttack}{https://github.com/hncszyq/ShadowAttack}.}.

\end{abstract}


\vspace{-0.1cm}
\section{Introduction}
\label{sec:intro}

In the past few years, we have witnessed the great success of deep neural networks (DNNs) in a variety of computer vision tasks, such as image classification, object detection, scene segmentation and so on. However, recent studies have revealed that DNNs based models are vulnerable to adversarial examples, even though the added magnitude of perturbations is small. In safety sensitive scenarios, such as autonomous driving\cite{Li_2019_CVPR} and medical diagnosis\cite{litjens2017survey}, adversarial inputs would enforce a machine learning system to produce erroneous decision, leading to unexpected situations that may be potentially dangerous.

Estimating when a machine learning model fails to work is of great concern to trustworthy AI. Accordingly, it is important to understand the level of risk of various adversarial examples to the machine learning models. There are numerous attack strategies investigated in the literature, which can be in the digital domain, where digital images corresponding to a scene are modified; or in the physical domain, where perturbations are physically added to the objects themselves \cite{Eykholt_2018_CVPR}.

The adversarial attacks in the physical domain receive more attention recently, since they are more practical and challenging. One popular approach for physical-world attacks is to adopt the ``sticker-pasting" strategy \cite{Eykholt_2018_CVPR}, where the adversarial perturbation is printed as a sticker and then attached onto the target object, \textit{e.g.}, a road sign. However, this approach would suffer from a few troubles: 1) In some cases, it may be impossible to access the target object; 2) The printing of perturbations is imperfect, \textit{i.e.}, some values cannot be reproduced by valid colors in the real world. Some strategies emerged very recently to remedy these limitations, which explored a new type of adversarial attack---the threats of light---to achieve non-invasive attack \cite{9156447,Nguyen_2020_CVPR_Workshops,laser,Gnanasambandam_2021_ICCV}. For instance, Duan \textit{et al.} \cite{laser} propose to utilize the laser beam as adversarial perturbation directly rather than crafting adversarial perturbation from scratch, which has been demonstrated to be an effective physical-world attack to DNNs.  Gnanasambandam \textit{et al.} \cite{Gnanasambandam_2021_ICCV} propose to project calculated patterns to alter the appearance of the target objects based on a low-cost projector-camera system. Experimental results shown in \cite{laser,Gnanasambandam_2021_ICCV} demonstrate the effectiveness of light attacks in both digital- and physical-settings. 

A basic principle in adversarial attack is that the carefully perturbed inputs should cause the network to generate wrong decision while without introducing a visible change to humans. However, the added optical patterns in existing methods, both laser beam \cite{laser} and projected pattern \cite{Gnanasambandam_2021_ICCV}, are artificial but not natural. Thus, they are still conspicuous and attention-grabbed, and can be easily noticed by humans.
In this paper, we study a new type of optical adversarial examples, in which the perturbations are generated by a very common natural phenomenon, \textit{shadow}, to achieve naturalistic and stealthy physical-world adversarial attack. We choose traffic sign recognition as our target task to validate the effectiveness of this attack. It is worth noting that, in the field of computer vision, many methods have been proposed to conduct shadow removal \cite{1542031,shadow_removal,shadow_tip}, which work from the perspective of image restoration in order to improve the accuracy of the subsequent high-level tasks \cite{shadow_face}. We are the first work to reveal that shadow can also become a threat to harm the reliability of machine learning based vision system. It is a meaningful reminding to prevent such attacks in practical systems.

The contributions of this paper can be highlighted as follows:
\begin{itemize}
    \item We propose a new light-based physical-world adversarial attack under the black-box setting via the very common natural phenomenon---shadow, which is more naturalistic and stealthy.
    \item We offer feasible optimization strategies to generate digitally and physically realizable adversarial examples perturbed by shadows.
    \item We provide thorough evaluations conducted on both simulated and real-world environments to demonstrate the effectiveness of our method. We also discuss the limitations and the defense mechanism of this attack.
\end{itemize}



\section{Related Work}
\label{sec:relatedwork}

\subsection{Adversarial Examples}

Adversarial examples were first discovered by Szegedy \etal  \cite{szegedy2013intriguing}, showing that a small perturbation can drastically change the network output while being quasi-imperceptible to humans. More interestingly, they also found the cross model generalization ability of adversarial examples, i.e., a large fraction of them will be misclassified by networks trained with different hyper-parameters or even with different architectures. This intriguing property of DNNs has greatly inspired the enthusiasm of academic research. Existing works on adversarial examples are either in the digital domain or in the physical world.

\subsection{Adversarial Examples in Digital Domain}

In the digital domain, adversarial perturbations are directly added to the inputs of a model, and their amplitudes are usually limited by $l_p$-norm, e.g. $l_{\infty}$-norm\cite{goodfellow2014explaining, madry2017towards}, $l_2$-norm and $l_0$-norm\cite{carlini2017towards} to ensure imperceptibility. According to adversary's knowledge, adversarial attacks can either be white-box, where an attacker knows all the details of a model so that the gradient can be fully used to craft a perturbation\cite{goodfellow2014explaining, kurakin2016adversarial, madry2017towards, carlini2017towards, papernot2016limitations}, or black-box, where an attacker can only query the target model and get a corresponding output without knowing its internal structure. In the black-box scenario, an attacker can take advantage of the cross model generalization ability of adversarial examples\cite{dong2018boosting, xie2019improving, dong2019evading, wu2020boosting} or reconstruct the internal information of a model through multiple queries\cite{chen2017zoo, ilyas2018black, brendel2017decision, chen2020hopskipjumpattack}. Further, these two strategies can be used at the same time to promote each other\cite{yang2020learning}.

\subsection{Adversarial Examples in Real Physical World}

Due to the variations caused by camera, a major concern is whether these highly optimized samples can pose a real threat to applications in the physical world. Kurakin \etal\cite{kurakin2016adversarial} first demonstrated the existence of physical adversarial examples. They printed out digital adversarial examples, and found that their corresponding photos taken by a mobile phone were still adversarial. However, such noise-based perturbations are not practical in the physical world, as they are hard to be captured by cameras at large distances. To make adversarial patterns more prominent, while keeping their stealthiness, a line of work hide perturbations with natural styles\cite{duan2020adversarial} or limit the large perturbations to a small area with specific semantics, e.g., stickers arranged in the shape of letters\cite{Eykholt_2018_CVPR} or a pair of glasses\cite{sharif2016accessorize}. To further improve robustness to the variations caused by camera, Athalye \etal\cite{athalye2018synthesizing} proposed Expectation Over Transformation (EOT), by which the first robust 3D adversarial example was generated.

Recently, a new line of work explores non-invasive attacks that do not access the target object, e.g., optic-based attacks. Sayles \etal\cite{sayles2021invisible} illuminates the target object with a special light that varies with a high frequency to cause rolling shutter effect of the camera. Duan \etal\cite{laser} disturb the classifier by generating a laser beam in front of the target object. Gnanasambandam \etal\cite{Gnanasambandam_2021_ICCV} project their calculated adversarial patterns to the target object by a projector. Instead of the optical phenomena produced by these sophisticated 
artificial devices, in this work, we turn to explore a common natural phenomenon---shadow.

\section{Approach}
\label{sec:shadowattack}

In this section, we present our method for learning an adversarial object by casting shadow to achieve successful attack. We first describe the representation of shadow.



\subsection{Problem Formulation}

Given an input image $x \in \mathbb R^d$ with the class label $y\in[1,\cdots,k]$, a DNN based classifier $f:  \mathbb R^d \rightarrow \mathbb{R}^k$ is trained to derive the predicted label $\widetilde{y}$:
\begin{equation}
    \widetilde{y}\triangleq\arg\max_if_i(x)
\end{equation}
where $f_i(x)$ is the confidence of the $i$-th class. 
The goal of our proposed method is to cast a specific shadow onto the target object to produce adversarial example $x_{adv}$, which causes the machine learning system to produce misclassification:
\begin{equation}
\label{eq:goal}
   \arg\max_if_i(x) \neq \arg\max_if_i(x_{adv})
\end{equation}
Meanwhile, it should guarantee that the perturbation imposed on $x_{adv}$ is imperceptible enough so that $x_{adv}$ is stealthy for humans.



\begin{figure}[t]
  \centering
  \includegraphics[width=1.0\linewidth]{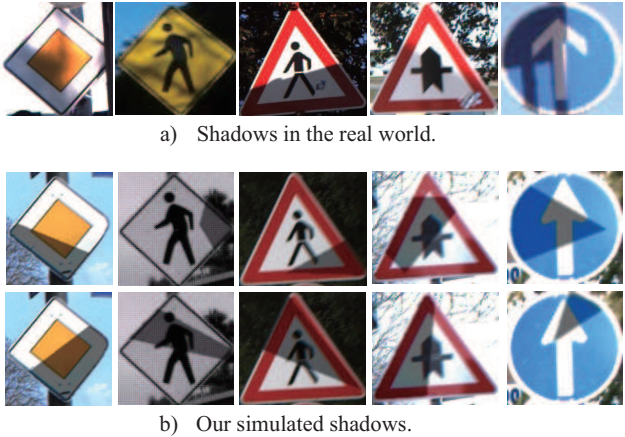}
  \vspace{-1.5em}
  \caption{Comparison of our simulated shadows and real-world shadows on traffic signs from LISA\cite{mogelmose2012vision} and GTSRB\cite{stallkamp2012man} datasets.}
  \label{fig:digitalexamples}
\end{figure}

\subsection{Shadow Perturbation Modeling}

To generate the adversarial example by shadow, two factors should be carefully considered: the location of shadow and the value of shadow.

\textbf{Shadow Location.} For the location of shadow, we propose to find an appropriate polygon $\mathcal{P_V}$ to simulate the shadow region, which is determined by a set of vertices $\mathcal{V}=\left\{(m_1,n_1),(m_2,n_2),...,(m_s,n_s)\right\}$. Given $\mathcal{M}$ as the mask to locate the target object, the shadow region in the target object can be defined as $\mathcal{M}\cap\mathcal{P_V}$. Note that, although $\mathcal{P_V}$ can be any valid polygon, based on the experimental experience shown in \cref{sec:experiments}, the simplest polygon---triangles---are sufficient to produce successful adversarial examples. We can exploit more complex polygon to achieve higher attack success rate, however, the shadow would look unnatural and it becomes more difficult to implement in the real world. Therefore, in practice, $\mathcal{P_V}$ is simply set to be a triangle. Thus, the optimization location parameters in our method is the corresponding vertices coordinates $\mathcal{V}=\{(m_1,n_1),(m_2,n_2),(m_3,n_3)\}$.

\textbf{Shadow Value.} To perform a successful physical-word attack, a main concern is the non-negligible gap between the digital and the physical domains. A widely used technique to tackle this problem is Expectation Over Transformation (EOT)\cite{athalye2018synthesizing}, which produces robust physical-world objects that remain adversarial when captured over a wide range of transformations, including distances, angles, exposures, etc. 

In addition to the variations caused by camera, in our context, another key factor should be considered is the relationship of pixel values between shadow and non-shadow areas in the real world scenario, since we need to simulate shadows in digital images when generating adversarial examples. However, the exact formula of this relationship is obscure. To simplify this problem, following the assumption in the field of shadow removal \cite{shadow_removal}, we regard that shadows only affect the illumination component and leave other ones remain unchanged. Accordingly, we transform images from RGB space to LAB space, and only consider the $L$ channel which is a correlate of lightness.
Our statistical results on SBU Shadow dataset \cite{vicente2016large} confirmed this assumption by indicating that the mean ratio of pixel values of channels $L$, $A$ and $B$ in shadow areas to that in non-shadow areas are 0.43, 0.99 and 0.90 respectively, and the corresponding standard deviations of channels $A$ and $B$ are as low as 0.05 and 0.07\footnote{We selected 400 images from the training set for statistical analysis.}. 

Based on the above analysis, we can generate shadow perturbation cast in an image by multiplying the corresponding $L$ channel in LAB color space with a coefficient $k$\footnote{Note that in addition to LAB color space, we have also tried YUV and HSL. Our statistical results show that LAB is the best choice.}. Specifically, given a clean image $x$ in RGB color space, we first convert $x$ to LAB color space:
\begin{equation}
    LAB(x)=LAB\left([R_x \ G_x \ B_x]\right)=[L_x \ A_x \ B_x]. \notag
\end{equation}
Then given a polygon $\mathcal{P_V}$ and a mask $\mathcal{M}$, the value of pixel $(i,j)$ in the shadow image $x_{adv}$ can be calculated as follows:
\begin{align}
    LAB^{ij}(x_{adv}) &= [L^{ij}_{x_{adv}} \ A^{ij}_{x_{adv}} \ B^{ij}_{x_{adv}}] \notag \\
                  &= \begin{cases}
                        LAB^{ij}(x) \cdot [k \ 1 \ 1]^T & (i,j)\in\mathcal{P_V}\cap\mathcal{M} \notag \\
                        LAB^{ij}(x) \cdot [1 \ 1 \ 1]^T & (i,j)\notin\mathcal{P_V}\cap\mathcal{M}
                     \end{cases}. \notag
\end{align}
Finally, we convert $x_{adv}$ back to RGB color space. For simplicity, we denote the above adversarial example generation process as:
\begin{equation}
    x_{adv}=\mathcal{S}(x, \mathcal{P_V}, \mathcal{M}, k)
    \label{eq:simulateShadow}
\end{equation}
In \cref{fig:digitalexamples}, we show some instances of our simulated shadows on traffic signs, as well as some real-world shadows for comparison.

\begin{figure}[t]
  \centering
  \includegraphics[width=1.0\linewidth]{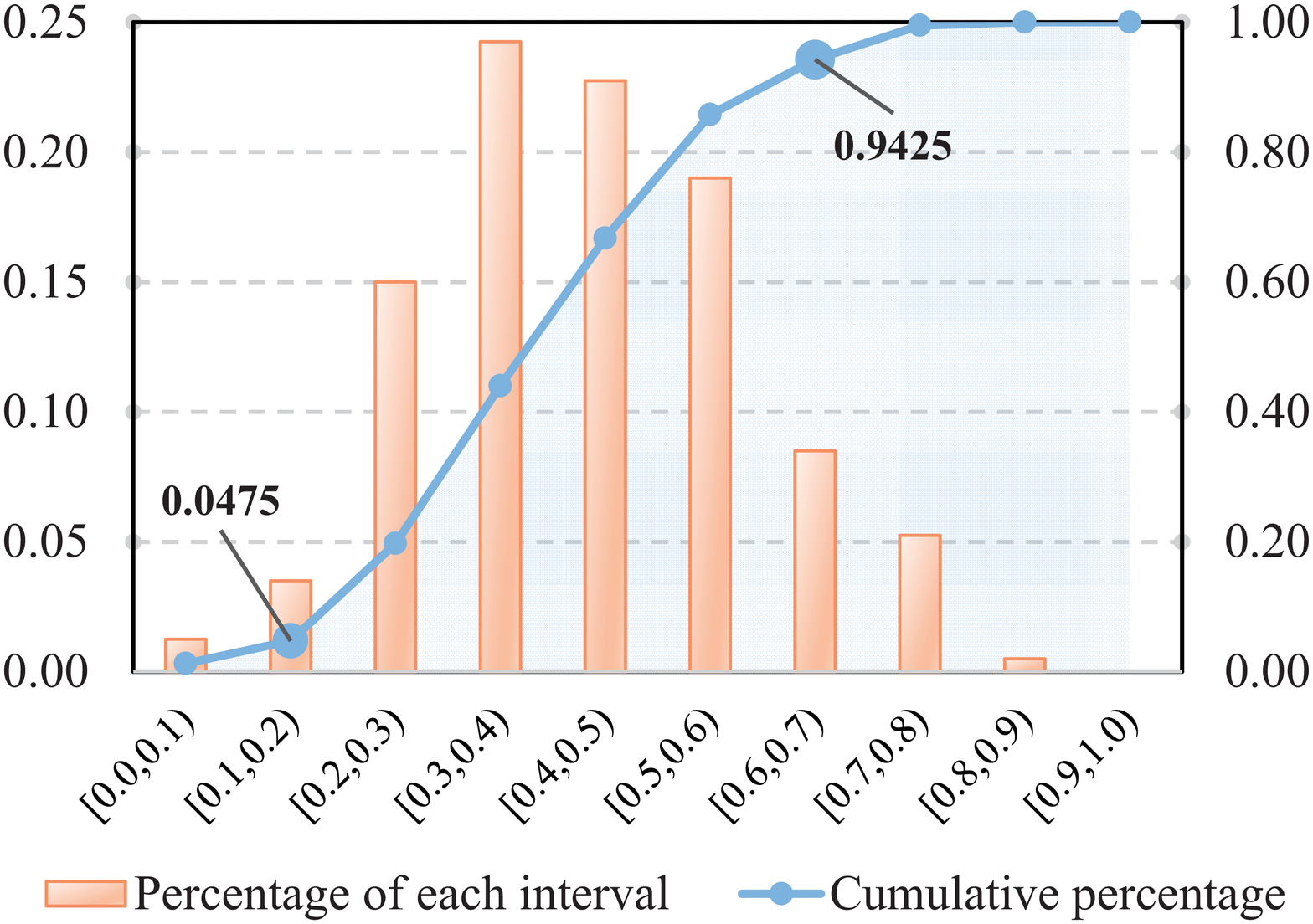}
  \vspace{-1.8em}
  \caption{\bf{The distribution of coefficient $k$}.}
  \label{fig:shadowstatistics}
\end{figure}

In practice, however, the form of shadow is the result of complex physical process. The coefficient $k$ can be varied due to numerous factors, including light sources, scene geometry, materials of the target object, and the imaging quality of the camera, etc. The distribution of $k$ in SBU Shadow dataset is shown in \cref{fig:shadowstatistics}. Accordingly, in our scheme, before generating an adversarial example, we first cast a random shadow on the target object to measure the coefficient $k$. Further, by leveraging EOT, the impact of the mismatch of $k$ can be minimized.

\subsection{Shadow Attack in Digital Domain}
\label{sec:digitaldomainattack}

In this subsection, we introduce in detail the generation process of adversarial examples using simulated shadows in the digital domain. 

The aim of our proposed attack method is to find vertices coordinates $\mathcal{V}=\{(m_1,n_1),(m_2,n_2),(m_3,n_3)\}$ of the triangle, which is used to simulate the shadow area, to make the resultant shadow image $x_{adv}$ being misclassified by the target model $f$. In this work,  we consider the more practical and challenging black-box attack scenario, where we can only access to the input (images) and the output (confidence scores $f(\cdot)$) of a targeted DNN, without knowing the detailed network structure and parameters. To this end, we formulate the optimization objective as minimizing the confidence score on true label:
\begin{equation}
    \mathop{\arg\min}\limits_{\mathcal{V}} \ f_{true}\left(\mathcal{S}(x, \mathcal{P_V}, \mathcal{M}, k)\right), \ \text{s.t.} \ \widetilde{y}_{adv}\neq y_{true}
    \label{eq:finalgoal}
\end{equation}
where $\widetilde{y}_{adv} = \arg\max_if_i(x_{adv})$.
We want to find the best set of coordinates $\mathcal{V}^*$ to minimize $f_{true}(x_{adv})$ so as to produce misclassification. 

In black-box scenarios, a popular approach is to use zeroth order optimization (ZOO) \cite{chen2017zoo} to estimate the gradients of the targeted DNN.
However, we observe severe gradient exploding and vanishing effect in the practical implementation of ZOO, which is probably due to: 1) the coordinate values in $\mathcal{V}$ are discrete, 2) the judgement on whether a coordinate locates within $\mathcal{P_V}\cap\mathcal{M}$ is a boolean function, making gradients unreliable. 

Instead, in our method, we turn to exploit the particle swarm optimization (PSO) strategy \cite{kennedy1995particle}, which stems from the study on how bird flocks search for food. Based on the cooperation and information sharing between individuals in a population, a valid solution can be found quickly without the help of gradients. Specifically, in PSO, we maintain a population of particles, each of which represents a candidate solution of \cref{eq:finalgoal} and has its own speed of movement in the solution space. In each iteration, all the particles adjust their speeds and positions according to the current individual optima and the global optimum shared by the whole population, and finally update every optimum. During this process, a cost function is needed to evaluate the quality of each candidate solution, which is the confidence score $f_{true}(\cdot)$ in our context. The algorithm terminates as long as $\widetilde{y}_{adv}\neq y_{true}$ or the maximum number of iterations is reached. To further increase the success rate, we adopt the $n$-random-restarts strategy, which means that when the attack fails we can reinitialize and rerun the PSO at most $n-1$ times.

\subsection{Shadow Attack in Real Physical-World}
\label{sec:physicaldomainattack}


\begin{table*}[t]
	\caption{Success rates of the proposed Shadow Attack with different shadow coefficient $k$ on LISA and GTSRB test sets. The column in bold refers to the performances when $k$ takes the mean of $k$ in SBU Shadow dataset, which is 0.43.}
	\vspace{-0.1cm}
	\centering
	\resizebox{\linewidth}{!}{
	\begin{tabular}{lcccccccccccc}
		\toprule
		\multirow{2}{*}{Model} 
		&\multicolumn{12}{c}{\textbf{Shadow Coefficient $k$}} \\
		
		\cline{2-13} 
		\specialrule{0em}{1pt}{1pt}
		&0.20 &0.25 &0.30 &0.35 &0.40 &\textbf{0.43} &0.45 &0.50 &0.55 &0.60 &0.65 &0.70\\
		\midrule
		LISA-CNN  
		&100.00 &100.00 &99.73 &99.18 &99.05 &\textbf{98.23} &97.95 &96.04 &93.18 &88.81 &83.08 &72.71 \\
		GTSRB-CNN          
		&97.37  &96.35  &95.14 &93.45 &91.36 &\textbf{90.47} &88.97 &87.15 &84.25 &80.36 &73.76 &66.73 \\
		\bottomrule
	\end{tabular}}
	\label{tab:digitalperformance}
	\vspace{-0.1cm}
\end{table*}

\begin{table*}[t]
	\caption{Average number of queries at the time when the attack succeeds with different shadow coefficient $k$ on LISA and GTSRB test sets. The column in bold refers to the performances when $k$ takes the mean of $k$ in SBU Shadow dataset, which is 0.43.}
	\vspace{-0.1cm}
	\centering
	\resizebox{\linewidth}{!}{
	\begin{tabular}{lcccccccccccc}
		\toprule
		\multirow{2}{*}{Model} 
		&\multicolumn{12}{c}{\textbf{Shadow Coefficient $k$}} \\
		
		\cline{2-13} 
		\specialrule{0em}{1pt}{1pt}
		&0.20 &0.25 &0.30 &0.35 &0.40 &\textbf{0.43} &0.45 &0.50 &0.55 &0.60 &0.65 &0.70\\
		\midrule
		LISA-CNN  
		&26.6 &27.7 &38.4 &71.8 &83.4 &\textbf{91.2} &100.6 &137.0 &169.2 &195.9 &306.7 &293.4 \\
		GTSRB-CNN          
		&98.3 &93.7 &112.8 &129.0 &128.4 &\textbf{126.8} &136.9 &155.9 &188.3 &232.0 &294.2 &343.4 \\
		\bottomrule
	\end{tabular}}
	\label{tab:digitalquerynum}
		\vspace{-0.3cm}
\end{table*}

To generate robust adversarial examples in the real physical-world, we adopt the following two strategies for enhancement:

\textbf{Expectation Over Transformation.}  EOT \cite{athalye2018synthesizing} is a powerful tool for dealing with the domain shift between the digital and the physical domains. To implement EOT, we start by defining a distribution of transformation $\mathcal{T}$ to simulate domain shift. Each instance of $\mathcal{T}$ is a combination of a set of random image transformations including down sampling, brightness adjustment, perspective transformation, motion blur. More importantly, the possible mismatch of shadow coefficient $k$ is also considered in $\mathcal{T}$ by setting a random $k$. Next, rather than optimizing $\mathcal{V}$ for a single image, we instead minimize the expectation of $f_{true}(\cdot)$ over the set of transformed images, \textit{i.e.}, 
\begin{equation}
    \mathop{\arg\min}\limits_{\mathcal{V}} \ \mathbb{E}_{t\sim\mathcal{T}}[f_{true}(t(x_{adv}))], \ \text{s.t.} \ \widetilde{y}_{adv}\neq y_{true}.
    \label{eq:eotgoal}
\end{equation}
In practice, we approximate the expectation in \cref{eq:eotgoal} by the mean of 10 transformed samples together with the original sample.

\textbf{Prediction stabilization}. According to Eykholt \etal \cite{Eykholt_2018_CVPR}, any successful physical perturbation must cause targeted misclassification in a range of distances and angles to avoid attack failure by performing simple
majority voting. However, due to the relatively small perturbation space, it is difficult to achieve targeted misclassification by shadows. Instead, we first optimize \cref{eq:eotgoal} such that the classifier should output a wrong decision $\widetilde{y}_w$. Then, we stabilize this prediction by rerunning PSO while conducting the following optimization:
\begin{equation}
    \mathop{\arg\max}\limits_{\mathcal{V}} \ \mathbb{E}_{t\sim\mathcal{T}}[f_w(t(x_{adv}))], \ \text{s.t.} \ \widetilde{y}_{adv}=\widetilde{y}_w.
    \label{eq:stabilizationobjective}
\end{equation}
Experiments in \cref{sec:physicaleval} demonstrate that our algorithm can achieve similar performance with targeted attacks in term of the stability of predictions.

\section{Experiments}
\label{sec:experiments}

\subsection{Datasets and Models}
We choose road sign classification as our target task. Following Eykholt \etal \cite{Eykholt_2018_CVPR}, we evaluate the performance of our algorithm on the same datasets and model architectures. One dataset is LISA \cite{mogelmose2012vision}, which consists of 47 different US road sign classes. To avoid the effect of long-tailed distribution, only the top 16 classes with the largest number of samples are tested.  The other dataset is GTSRB \cite{stallkamp2012man}, which consists of 43 different German road sign classes. 
Note that there are some images in both datasets that are captured under very dark condition or already in shadows. In view of this, we remove images whose average pixel value of L-channel in the area of traffic sign is less than 120.

\subsection{Evaluation in Digital Domain}

In the digital domain, we attack every image in the test sets of LISA and GTSRB with simulated shadow. We measure the performance of our algorithm by attack success rate, which is defined as the ratio of the number of samples that successfully cause misclassification to the number of all samples in the test set. As shown in \cref{fig:shadowstatistics}, the shadow value $k$ can be varied in a wide range in the real world. Therefore, we repeat the experiment with several different values of $k$ ranging from 0.2 to 0.7, and the results are shown in \cref{tab:digitalperformance}. It can be found that, when $k$ is set to 0.43, which is the mean of shadow values in the SBU Shadow dataset \cite{vicente2016large}, our success rates on LISA and GTSRB test sets are 98.23\% and 90.47\% respectively. When $k>0.7$, the successful rates are relatively low, because the shadows are weak and the perturbed images are close to the clean ones. 

In the black-box scenario, to keep stealthy, another important concern is how many queries should be performed. We investigate this point in \cref{tab:digitalquerynum}, from which it can be found that our method can get a valid adversarial example with only dozens or hundreds of queries.

\subsection{Evaluation in Physical Domain}
\label{sec:physicaleval}

In the physical domain, to simulate real-world application scenarios, we first cast the specific shadow onto our target traffic sign and then record a video with a camera moving towards the sign. Next, we measure the performance of our attack by the percentage of misclassified frames in this video. By experiments, we demonstrate that our method can consistently carry out effective attacks not only outdoors in the daytime using sunlight to generate shadows, but also at night or indoors by leveraging artificial light source to generate shadows.

\begin{figure*}[t]
  \vspace{-0.8em}
  \centering
  \includegraphics[width=1.0\linewidth]{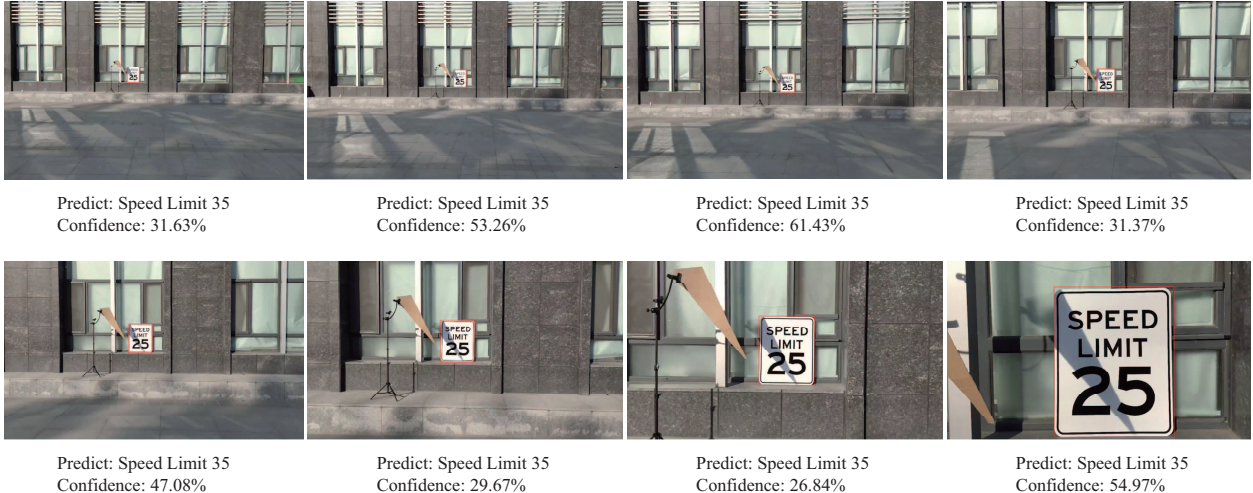}
  \vspace{-1.5em}
  \caption{Examples of frames in the video of our outdoor experiment and the corresponding classification results.}
  \label{fig:physicalinstances}
  \vspace{-0.3cm}
\end{figure*}

\textbf{Outdoor Environment.} We take the US speed limit 25 sign as an example for outdoor test in the daytime. As shown in \cref{fig:physicalinstances}, we generate the shadow by a cardboard. We record a video with 220 frames to simulate a self-driving car coming from far to near. For each frame, we first manually draw the bounding-box of the traffic sign and then follow the crop-resize-then-classify pipeline for classification. The results show that, our attack achieves 100.00\% misclassification rate, while 95.91\% of the frames are misclassified as speed limit 35. This 
implies that the performance of our untargeted attack is comparable to that of targeted attacks in term of the stability of predictions, making the voting algorithm fail to be an effective defense. \cref{fig:physicalinstances} shows some frames at equal intervals of time as well as their predictions.

\begin{figure*}[t]
  \centering
  \includegraphics[width=0.95\linewidth]{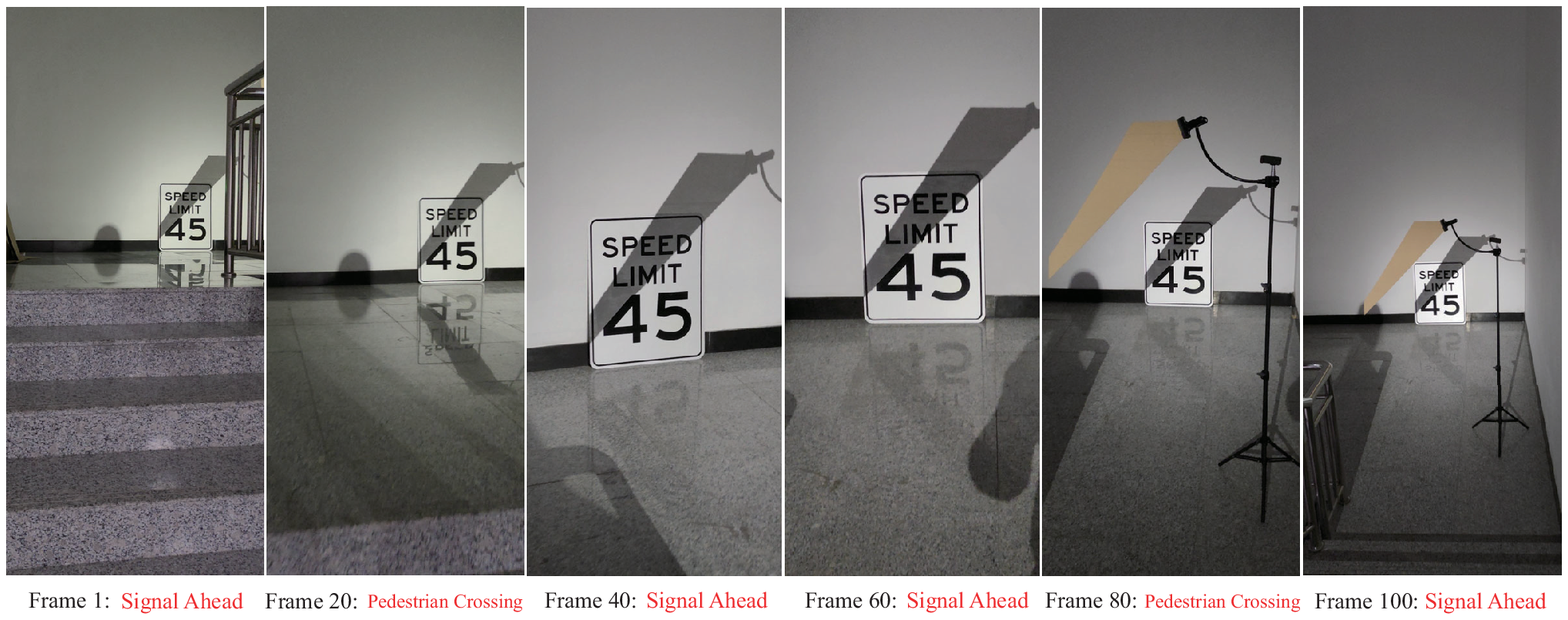}
  \caption{Examples of frames in a video of our indoor experiment and the corresponding classification results.}
  \label{fig:indoortest}
\end{figure*}

\textbf{Indoor Environment.} One limitation of our attack is that it is difficult to produce obvious shadows at night without sunlight or indoors with poor lighting. In this case, we use a flashlight to generate artificial light, which achieves comparable effect to sunlight. Our indoor test is conducted in a dark stairwell, as illustrated in \cref{fig:indoortest}. The traffic signs are placed on the platform in the middle of two floors and the flashlight is fixed on the upper floor while ensuring that the light can be cast onto the traffic signs. We take four US speed limit signs from LISA \cite{mogelmose2012vision} as examples for indoor test. For each sign, we record a video  when walking from the lower floor to the upper floor. The videos include 100 frames. \cref{tab:indoortest} offers the misclassification rate of each video, which demonstrates that our attack is also effective indoors. \cref{fig:indoortest} provides some example frames in the test of the speed limit 45 sign, in which their predictions are also given. It can be found our attack is successful for all.

\begin{table}[t]
	\caption{Experimental results of our indoor test, where error rate refers to the percentage of misclassified frames in each video and stability refers to the percentage of frames predicted as the primary error class.}
	\centering
	\resizebox{\linewidth}{!}{
	\begin{tabular}{lccc}
		\toprule
		Ground truth &\textbf{Predict} &\textbf{Error rate} &\textbf{Stability} \\
		
		\midrule
		Speed limit 25 &Speed limit 35 &100\% &88\% \\
		Speed limit 30 &Speed limit 35 &100\% &93\% \\
		Speed limit 35 &Speed limit 30 &95\%  &71\% \\
		Speed limit 45 &Signal Ahead   &100\% &76\% \\

		\bottomrule

	\end{tabular}}
	\label{tab:indoortest}
	\vspace{-0.5cm}
\end{table}

\subsection{Scheduled Attack}

Once the cardboard is fixed, the generated shadow will move over time due to the angle change of sunlight, which inspires us that our method can be used to conduct an interesting scheduled attack. Scheduled attack means that, due to the movement of the sun, we can control our shadow perturbation to be harmless in most of the time but to be adversarial at a certain time, e.g. the peak traffic hour in the morning. 

To carry out a scheduled attack, we just need to fix the cardboard as done in \cref{sec:physicaleval} at a specific time of a day in advance, and then the adversarial effect would be generated at the same time the future days if there is no significant change of the weather. Furthermore, after applying the strategies mentioned in \cref{sec:physicaldomainattack} to enhance robustness, our attack can maintain successful for a period of time, rather than only effective at a certain moment.

To test the scheduled attack, we simulate it on digital images. We first place our targeted traffic sign image on the XOZ plane of a three-dimensional coordinate system. Next, we calculate the solar elevation angle and the solar azimuth angle by using our scheduled time, longitude and latitude. Based on this, as long as we set the distance to the traffic sign in the Y-axis direction, we can further get the coordinates of each vertex of the cardboard. By fixing these coordinates, we can figure out the positions of shadows on the XOZ plane at any time. In this way, we simulate the dynamic changes of shadows over time.

\begin{figure*}[t]
  \centering
  \vspace{-0.2cm}
  \includegraphics[width=0.95\linewidth]{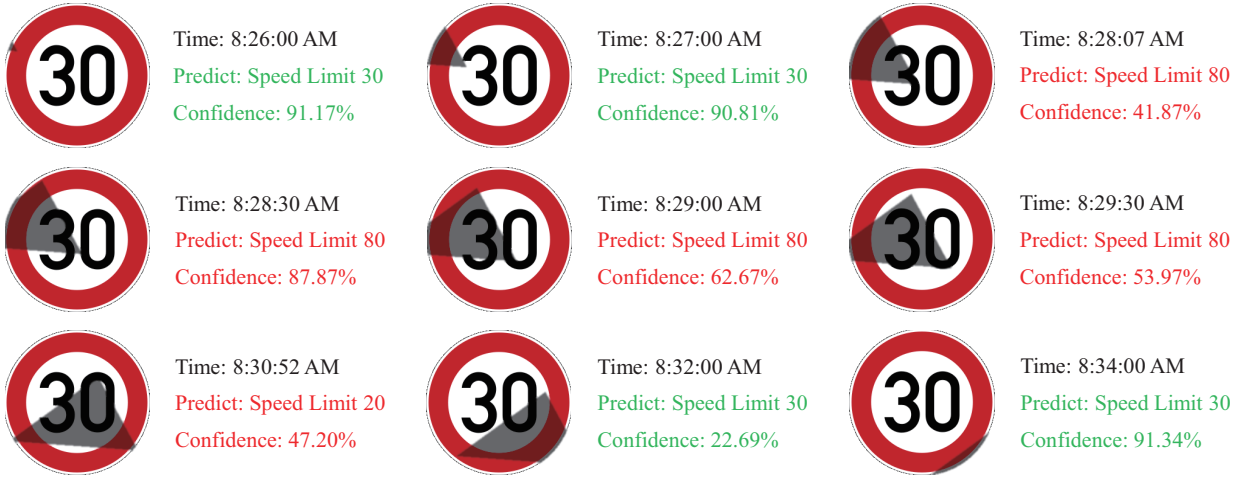}
  \caption{Our simulated scheduled attack. Our scheduled time is 8:30 in the morning, and the target class is speed limit 80. We show how the predictions change as the shadow moves on the German speed limit 30 sign. The predictions before 8:28:07 am and after 8:30:52 am are correct, while the predictions from 8:28:07 am to 8:30:06 am are speed limit 80 and the predictions from 8:30:07 am to 8:30:52 am are speed limit 20.}
  \label{fig:scheduledattack}
  \vspace{-0.2cm}
\end{figure*}

\begin{figure}[t]
  \centering
  \includegraphics[width=1.0\linewidth,trim=105 5 105 5]{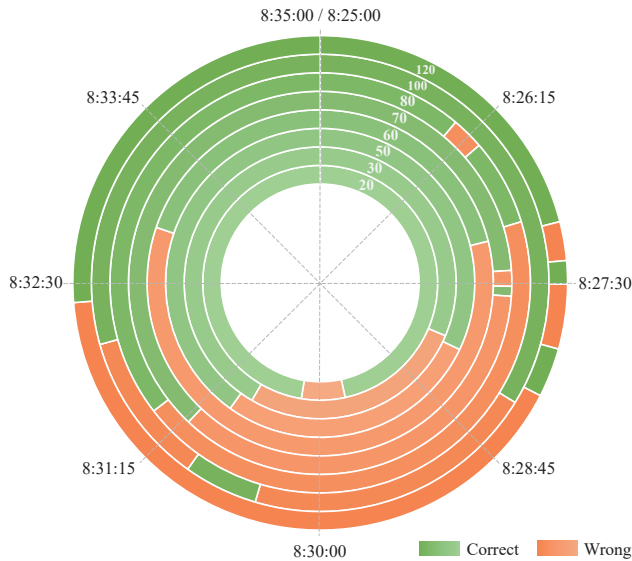}
  \caption{The test result of our scheduled attack on 8 German speed limit signs. The green and red areas represent correct and incorrect classifications respectively.}
  \label{fig:schedulestatistic}
  	\vspace{-0.3cm}
\end{figure}

We choose eight German speed limit signs from GTSRB \cite{stallkamp2012man} as test examples for our experiment. The longitude and latitude are set to 0 and 45 respectively. The traffic signs face south, i.e., the Y-axis points to the south. The distances from cardboard to traffic signs in the Y-axis direction are set to one meter. For each sign, we set 8:30 in the morning as the scheduled time, and we generate images with shadow every second from 8:25 to 8:35, from which we can observe how the classifier's predictions change over time. We show an instance of our simulated scheduled attack in \cref{fig:scheduledattack}. As shown in \cref{fig:schedulestatistic}, we achieve the goal of scheduled attacks on all eight speed limit signs. The duration of successful attack ranges from dozens of seconds to several minutes. 

\subsection{Ablation study}

\begin{table}[t]
	\caption{Performances of the proposed Shadow Attack with different number of edges of polygon $\mathcal{P_V}$.}
	\centering
	\begin{tabular}{lcccc}
		\toprule
		\multirow{2}{*}{Model} 
		&\multicolumn{4}{c}{\textbf{Number of edges}} \\
		
		\cline{2-5} 
		\specialrule{0em}{1pt}{1pt}
		&3 &5 &7 &9 \\
		\midrule
		LISA-CNN
		&97.95 &98.91 &99.45 &99.59 \\
		GTSRB-CNN
		&90.80 &93.72 &96.83 &97.59 \\
		\bottomrule
	\end{tabular}
	\label{tab:polygonablation}
	\vspace{-0.3cm}
\end{table}

\begin{table}[t]
	\caption{ablation study of $n$-random-starts strategy.}
	\vspace{-0.5em}
	\centering
	\resizebox{\linewidth}{!}{
	\begin{tabular}{lcccccc}
		\toprule
		\multirow{2}{*}{Model} 
		&\multicolumn{5}{c}{\textbf{Number of restarts}} \\
		
		\cline{2-6} 
		\specialrule{0em}{1pt}{1pt}
		&1 &5 &10 &50 &100 \\
		\midrule
		LISA-CNN
		&95.91 &98.36 &98.50 &99.05 &99.18 \\
		GTSRB-CNN
		&87.76 &90.74 &91.76 &92.81 &93.22 \\
		\bottomrule
	\end{tabular}}
	\label{tab:randomstartablation}
	\vspace{-0.3cm}
\end{table}

\textbf{About the number of edges of $\bm{\mathcal{P_V}}$}. Theoretically, $\mathcal{P_V}$ can be any valid polygon. However, when the number of edges is relatively large, the shadow would look unnatural, which goes against the requirement of stealthiness. Moreover, a polygon with too many edges leads to more computational overhead and it is not easy to implement in the real world. So in practice, we choose to attack the target object by a simple polygon area. Indeed, as shown in \cref{tab:polygonablation}, the simplest polygon---triangles---are sufficient to produce successful adversarial examples. The performance loss caused by the reduction in the number of edges is also acceptable. 

\textbf{About $\bm{n}$-random-restarts strategy}. We test the effectiveness of $n$-random-restarts by setting $n$ to 1, 5, 10, 50, and 100 respectively. As shown in \cref{tab:randomstartablation}, compared with $n=1$, by setting $n=5$  we can effectively improve the success rate. However, when $n$ is greater than 5, there are only marginal improvements in performance by increasing $n$, demonstrating the effectiveness of PSO. To balance the success rate and computational overhead, we set $n=5$ in practice.

\subsection{How to Defend against the Shadow Attack?}

\begin{table}[t]
	\caption{Performance comparison of models with and without robustness training, where accuracy denotes the clean accuracy in LISA and GTSRB test sets, robustness denotes the probability that our attack fails (1 - attack success rate), and queries denotes the average number of queries at the time when our attack succeeds. All the data are obtained when $k=0.43$.}
	\vspace{-0.1cm}
	\centering
	\resizebox{\linewidth}{!}{
	\begin{tabular}{lcccc}
		\toprule
		Model &\textbf{Accuracy} &\textbf{Robustness} &\textbf{Queries} \\
		
		\midrule
		LISA-CNN
		&99.63 &1.73 &91.24 \\
		GTSRB-CNN
		&99.00 &9.53 &126.75 \\
		
		\midrule
		$\text{LISA-CNN}_{rob}$
		&99.56 &40.93 &849.51 \\
		$\text{GTSRB-CNN}_{rob}$
		&98.91 &25.57 &464.52 \\
		\bottomrule
	\end{tabular}}
	\label{tab:robustmodelresult}
		\vspace{-0.3cm}
\end{table}

In the above subsections, we demonstrate that the shadow perturbations is able to pose a security threat to the traffic or other security-critical scenarios. We discuss the defense mechanism against this type of attack as follows. 

One of the most effective defense strategies against adversarial examples is adversarial training (AT)\cite{madry2017towards}, which improves model robustness by using newly generated adversarial examples as training samples in each epoch. In our context, to implement AT, we need to apply the algorithm described in \cref{sec:digitaldomainattack} to each training sample in an epoch, which however is a particularly time-consuming process. In order to make training faster,  in each epoch we only add a random shadow to each training sample (the coordinate values in $\mathcal{V}$ and the coefficient $k$ are random variables), instead of the worst case shadow. We expect that such a training method can reduce the model's sensitivity to shadows, thereby improving the robustness to our attack. We retrain the models in this way, which are denoted as $\text{LISA-CNN}_{rob}$ and $\text{GTSRB-CNN}_{rob}$. To verify the effectiveness of the modified adversarial training, we perform attack on the retrained models in the digital domain. The results summarized in \cref{tab:robustmodelresult} demonstrate that this defense can improve the model robustness and the difficulty of the attack significantly at the expense of slightly decreased accuracy on clean samples.

\section{Limitation}

One limitation of our attack is the requirement of a strong single light source. In the environment with poor lighting condition, our method is difficult to produce significant shadow perturbations, \textit{i.e.}, $k$ would be large, which result in little difference between the attacked images and the clean ones. As shown in \cref{tab:digitalperformance} and \cref{tab:digitalquerynum}, a large $k$ leads to a low success rate and more queries. To remedy this problem, we can achieve the adversarial effect through a stronger artificial light as shown in \cref{sec:physicaleval}. For the scenarios with multiple light sources, it is also difficult for our method to produce effective shadow perturbations. 

Another limitation is that we cannot explicitly conduct targeted attacks, which is probably due to the perturbation directions produced by shadows are relatively single. However, as shown in our outdoor experiment, we are able to achieve comparable results to that of targeted attacks in term of the stability of predictions.

\section{Conclusion}
\label{sec:conclusion}

In this paper, we present a new optics-based adversarial attack strategy, which utilizes the very common natural phenomenon, shadow, to produce harmful perturbations on the targets. Experimental results demonstrate this attack can be successfully applied in both digital- and physical-settings.
Our work reveals that shadows can be dangerous, which is able to mislead the machine learning based vision system to produce erroneous decision.

As a new type of attack, there surely would be some drawbacks or limitations. We will work on making it more practical in future research. 

\section{Acknowledgement}

This work was supported by  National Natural Science Foundation of China under Grants 61922027, 6207115 and 61932022.

{\small
\bibliographystyle{ieee_fullname}
\bibliography{ShadowAttack}

\begin{thebibliography}{10}\itemsep=-1pt

\bibitem{athalye2018synthesizing}
Anish Athalye, Logan Engstrom, Andrew Ilyas, and Kevin Kwok.
\newblock Synthesizing robust adversarial examples.
\newblock In {\em International conference on machine learning}, pages
  284--293. PMLR, 2018.

\bibitem{brendel2017decision}
Wieland Brendel, Jonas Rauber, and Matthias Bethge.
\newblock Decision-based adversarial attacks: Reliable attacks against
  black-box machine learning models.
\newblock {\em arXiv preprint arXiv:1712.04248}, 2017.

\bibitem{carlini2017towards}
Nicholas Carlini and David Wagner.
\newblock Towards evaluating the robustness of neural networks.
\newblock In {\em 2017 ieee symposium on security and privacy (sp)}, pages
  39--57. IEEE, 2017.

\bibitem{chen2020hopskipjumpattack}
Jianbo Chen, Michael~I Jordan, and Martin~J Wainwright.
\newblock Hopskipjumpattack: A query-efficient decision-based attack.
\newblock In {\em 2020 ieee symposium on security and privacy (sp)}, pages
  1277--1294. IEEE, 2020.

\bibitem{chen2017zoo}
Pin-Yu Chen, Huan Zhang, Yash Sharma, Jinfeng Yi, and Cho-Jui Hsieh.
\newblock Zoo: Zeroth order optimization based black-box attacks to deep neural
  networks without training substitute models.
\newblock In {\em Proceedings of the 10th ACM workshop on artificial
  intelligence and security}, pages 15--26, 2017.

\bibitem{dong2018boosting}
Yinpeng Dong, Fangzhou Liao, Tianyu Pang, Hang Su, Jun Zhu, Xiaolin Hu, and
  Jianguo Li.
\newblock Boosting adversarial attacks with momentum.
\newblock In {\em Proceedings of the IEEE conference on computer vision and
  pattern recognition}, pages 9185--9193, 2018.

\bibitem{dong2019evading}
Yinpeng Dong, Tianyu Pang, Hang Su, and Jun Zhu.
\newblock Evading defenses to transferable adversarial examples by
  translation-invariant attacks.
\newblock In {\em Proceedings of the IEEE/CVF Conference on Computer Vision and
  Pattern Recognition}, pages 4312--4321, 2019.

\bibitem{duan2020adversarial}
Ranjie Duan, Xingjun Ma, Yisen Wang, James Bailey, A~Kai Qin, and Yun Yang.
\newblock Adversarial camouflage: Hiding physical-world attacks with natural
  styles.
\newblock In {\em Proceedings of the IEEE/CVF conference on computer vision and
  pattern recognition}, pages 1000--1008, 2020.

\bibitem{laser}
Ranjie Duan, Xiaofeng Mao, A.~K. Qin, Yuefeng Chen, Shaokai Ye, Yuan He, and
  Yun Yang.
\newblock Adversarial laser beam: Effective physical-world attack to dnns in a
  blink.
\newblock In {\em 2021 IEEE/CVF Conference on Computer Vision and Pattern
  Recognition (CVPR)}, pages 16057--16066, 2021.

\bibitem{Eykholt_2018_CVPR}
Kevin Eykholt, Ivan Evtimov, Earlence Fernandes, Bo Li, Amir Rahmati, Chaowei
  Xiao, Atul Prakash, Tadayoshi Kohno, and Dawn Song.
\newblock Robust physical-world attacks on deep learning visual classification.
\newblock In {\em Proceedings of the IEEE Conference on Computer Vision and
  Pattern Recognition (CVPR)}, June 2018.

\bibitem{1542031}
G.D. Finlayson, S.D. Hordley, Cheng Lu, and M.S. Drew.
\newblock On the removal of shadows from images.
\newblock {\em IEEE Transactions on Pattern Analysis and Machine Intelligence},
  28(1):59--68, 2006.

\bibitem{Gnanasambandam_2021_ICCV}
Abhiram Gnanasambandam, Alex~M. Sherman, and Stanley~H. Chan.
\newblock Optical adversarial attack.
\newblock In {\em Proceedings of the IEEE/CVF International Conference on
  Computer Vision (ICCV) Workshops}, pages 92--101, October 2021.

\bibitem{goodfellow2014explaining}
Ian~J Goodfellow, Jonathon Shlens, and Christian Szegedy.
\newblock Explaining and harnessing adversarial examples.
\newblock {\em arXiv preprint arXiv:1412.6572}, 2014.

\bibitem{ilyas2018black}
Andrew Ilyas, Logan Engstrom, Anish Athalye, and Jessy Lin.
\newblock Black-box adversarial attacks with limited queries and information.
\newblock In {\em International Conference on Machine Learning}, pages
  2137--2146. PMLR, 2018.

\bibitem{kennedy1995particle}
James Kennedy and Russell Eberhart.
\newblock Particle swarm optimization.
\newblock In {\em Proceedings of ICNN'95-international conference on neural
  networks}, volume~4, pages 1942--1948. IEEE, 1995.

\bibitem{kurakin2016adversarial}
Alexey Kurakin, Ian Goodfellow, Samy Bengio, et~al.
\newblock Adversarial examples in the physical world, 2016.

\bibitem{shadow_removal}
Hieu Le and Dimitris Samaras.
\newblock Physics-based shadow image decomposition for shadow removal.
\newblock {\em IEEE Transactions on Pattern Analysis and Machine Intelligence},
  pages 1--1, 2021.

\bibitem{Li_2019_CVPR}
Peiliang Li, Xiaozhi Chen, and Shaojie Shen.
\newblock Stereo r-cnn based 3d object detection for autonomous driving.
\newblock In {\em Proceedings of the IEEE/CVF Conference on Computer Vision and
  Pattern Recognition (CVPR)}, June 2019.

\bibitem{litjens2017survey}
Geert Litjens, Thijs Kooi, Babak~Ehteshami Bejnordi, Arnaud Arindra~Adiyoso
  Setio, Francesco Ciompi, Mohsen Ghafoorian, Jeroen~Awm Van Der~Laak, Bram
  Van~Ginneken, and Clara~I S{\'a}nchez.
\newblock A survey on deep learning in medical image analysis.
\newblock {\em Medical image analysis}, 42:60--88, 2017.

\bibitem{madry2017towards}
Aleksander Madry, Aleksandar Makelov, Ludwig Schmidt, Dimitris Tsipras, and
  Adrian Vladu.
\newblock Towards deep learning models resistant to adversarial attacks.
\newblock {\em arXiv preprint arXiv:1706.06083}, 2017.

\bibitem{mogelmose2012vision}
Andreas Mogelmose, Mohan~Manubhai Trivedi, and Thomas~B Moeslund.
\newblock Vision-based traffic sign detection and analysis for intelligent
  driver assistance systems: Perspectives and survey.
\newblock {\em IEEE Transactions on Intelligent Transportation Systems},
  13(4):1484--1497, 2012.

\bibitem{Nguyen_2020_CVPR_Workshops}
Dinh-Luan Nguyen, Sunpreet~S. Arora, Yuhang Wu, and Hao Yang.
\newblock Adversarial light projection attacks on face recognition systems: A
  feasibility study.
\newblock In {\em Proceedings of the IEEE/CVF Conference on Computer Vision and
  Pattern Recognition (CVPR) Workshops}, June 2020.

\bibitem{papernot2016limitations}
Nicolas Papernot, Patrick McDaniel, Somesh Jha, Matt Fredrikson, Z~Berkay
  Celik, and Ananthram Swami.
\newblock The limitations of deep learning in adversarial settings.
\newblock In {\em 2016 IEEE European symposium on security and privacy
  (EuroS\&P)}, pages 372--387. IEEE, 2016.

\bibitem{sayles2021invisible}
Athena Sayles, Ashish Hooda, Mohit Gupta, Rahul Chatterjee, and Earlence
  Fernandes.
\newblock Invisible perturbations: Physical adversarial examples exploiting the
  rolling shutter effect.
\newblock In {\em Proceedings of the IEEE/CVF Conference on Computer Vision and
  Pattern Recognition}, pages 14666--14675, 2021.

\bibitem{sharif2016accessorize}
Mahmood Sharif, Sruti Bhagavatula, Lujo Bauer, and Michael~K Reiter.
\newblock Accessorize to a crime: Real and stealthy attacks on state-of-the-art
  face recognition.
\newblock In {\em Proceedings of the 2016 acm sigsac conference on computer and
  communications security}, pages 1528--1540, 2016.

\bibitem{stallkamp2012man}
Johannes Stallkamp, Marc Schlipsing, Jan Salmen, and Christian Igel.
\newblock Man vs. computer: Benchmarking machine learning algorithms for
  traffic sign recognition.
\newblock {\em Neural networks}, 32:323--332, 2012.

\bibitem{szegedy2013intriguing}
Christian Szegedy, Wojciech Zaremba, Ilya Sutskever, Joan Bruna, Dumitru Erhan,
  Ian Goodfellow, and Rob Fergus.
\newblock Intriguing properties of neural networks.
\newblock {\em arXiv preprint arXiv:1312.6199}, 2013.

\bibitem{9156447}
James Tu, Mengye Ren, Sivabalan Manivasagam, Ming Liang, Bin Yang, Richard Du,
  Frank Cheng, and Raquel Urtasun.
\newblock Physically realizable adversarial examples for lidar object
  detection.
\newblock In {\em 2020 IEEE/CVF Conference on Computer Vision and Pattern
  Recognition (CVPR)}, pages 13713--13722, 2020.

\bibitem{vicente2016large}
Tom{\'a}s F~Yago Vicente, Le Hou, Chen-Ping Yu, Minh Hoai, and Dimitris
  Samaras.
\newblock Large-scale training of shadow detectors with noisily-annotated
  shadow examples.
\newblock In {\em European Conference on Computer Vision}, pages 816--832.
  Springer, 2016.

\bibitem{wu2020boosting}
Weibin Wu, Yuxin Su, Xixian Chen, Shenglin Zhao, Irwin King, Michael~R Lyu, and
  Yu-Wing Tai.
\newblock Boosting the transferability of adversarial samples via attention.
\newblock In {\em Proceedings of the IEEE/CVF Conference on Computer Vision and
  Pattern Recognition}, pages 1161--1170, 2020.

\bibitem{xie2019improving}
Cihang Xie, Zhishuai Zhang, Yuyin Zhou, Song Bai, Jianyu Wang, Zhou Ren, and
  Alan~L Yuille.
\newblock Improving transferability of adversarial examples with input
  diversity.
\newblock In {\em Proceedings of the IEEE/CVF Conference on Computer Vision and
  Pattern Recognition}, pages 2730--2739, 2019.

\bibitem{yang2020learning}
Jiancheng Yang, Yangzhou Jiang, Xiaoyang Huang, Bingbing Ni, and Chenglong
  Zhao.
\newblock Learning black-box attackers with transferable priors and query
  feedback.
\newblock {\em Advances in Neural Information Processing Systems}, 33, 2020.

\bibitem{shadow_tip}
Ling Zhang, Qing Zhang, and Chunxia Xiao.
\newblock Shadow remover: Image shadow removal based on illumination recovering
  optimization.
\newblock {\em IEEE Transactions on Image Processing}, 24(11):4623--4636, 2015.

\bibitem{shadow_face}
Wuming Zhang, Xi Zhao, Jean-Marie Morvan, and Liming Chen.
\newblock Improving shadow suppression for illumination robust face
  recognition.
\newblock {\em IEEE Transactions on Pattern Analysis and Machine Intelligence},
  41(3):611--624, 2019.

\end{thebibliography}
}

\end{document}